\documentclass[10pt,twocolumn,letterpaper]{article}

\usepackage{iccv}
\usepackage{times}
\usepackage{epsfig}
\usepackage{graphicx}
\usepackage{array, multirow}
\usepackage{amsmath}
\usepackage{amssymb}
\usepackage{caption}
\usepackage{subcaption}
\usepackage{comment}
\usepackage{bbm}
\usepackage[noend]{algpseudocode}
\usepackage{algpseudocode}
\usepackage{algorithm}
\usepackage[numbers,sort]{natbib}

\usepackage[breaklinks=true,bookmarks=false]{hyperref}

\iccvfinalcopy 

\def\iccvPaperID{1449} 

\ificcvfinal\fi
\begin{document}
	
	\title{Multi-Cue Structure Preserving MRF for Unconstrained Video Segmentation}
	
	\author{Saehoon Yi and Vladimir Pavlovic\\
		Rutgers, The State University of New Jersey\\
		110 Frelinghuysen Road, Piscataway, NJ 08854, USA\\
		{\tt\small \{shyi, vladimir\}@cs.rutgers.edu}
	}
	
	\maketitle

	\begin{abstract}
		Video segmentation is a stepping stone to understanding video context.  Video segmentation enables one to represent a video by decomposing it into coherent regions which comprise whole or parts of objects.  However, the challenge originates from the fact that most of the video segmentation algorithms are based on unsupervised learning due to expensive cost of pixelwise video annotation and intra-class variability within similar unconstrained video classes.
		We propose a Markov Random Field model for unconstrained video segmentation that relies on tight integration of multiple cues: vertices are defined from contour based superpixels, unary potentials from temporal smooth label likelihood and pairwise potentials from global structure of a video.  Multi-cue structure is a breakthrough to extracting coherent object regions for unconstrained videos in absence of supervision.  Our experiments on VSB100 dataset show that the proposed model significantly outperforms competing state-of-the-art algorithms.  Qualitative analysis illustrates that video segmentation result of the proposed model is consistent with human perception of objects.
	\end{abstract}
	
	\section{Introduction}
	
	\begin{figure*}[ht]
		\centering
		\includegraphics[width=\textwidth]{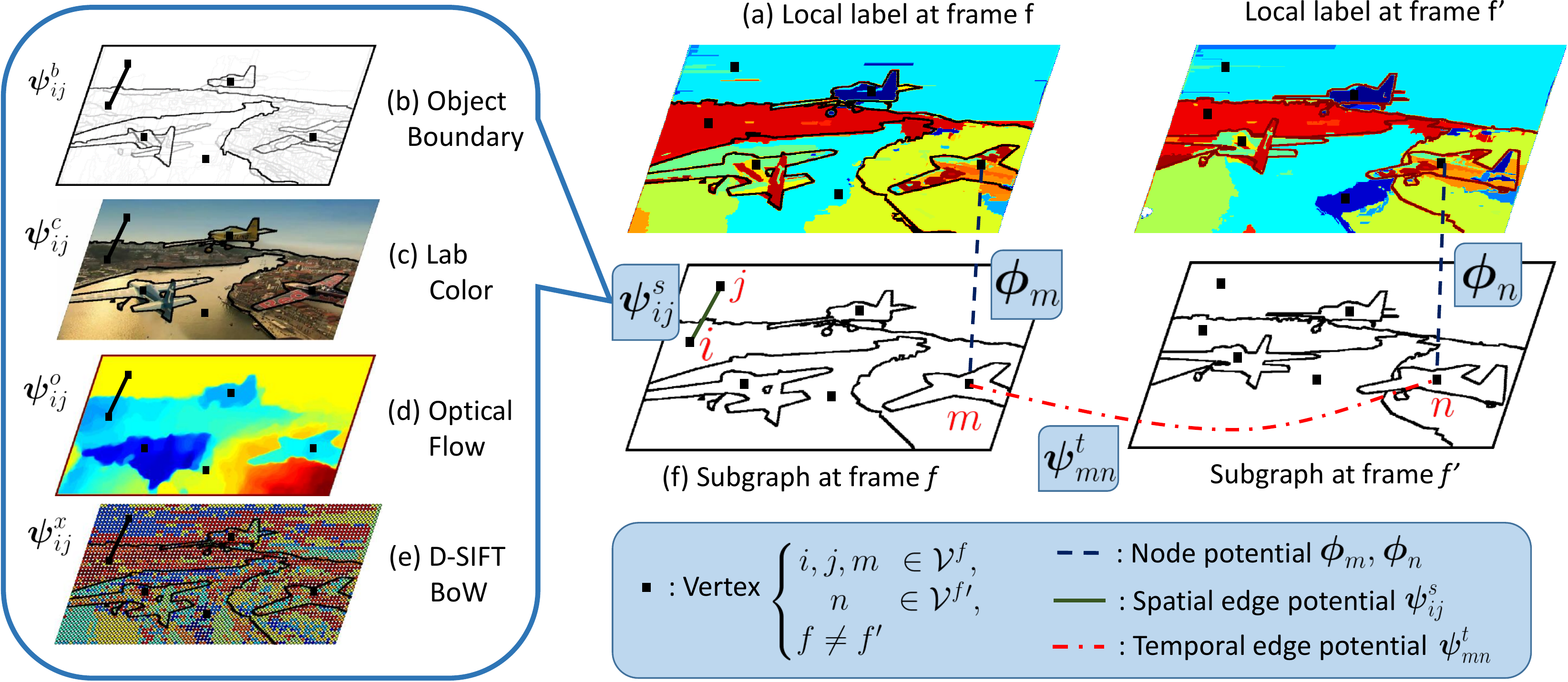}
		\caption{Overview of the framework. 
			(a) Node potential depends on histogram of temporal smooth pixelwise labels of the corresponding frame.  Spatial edge potentials: (b) Gray intensity represents contour strength.  (c) RGB color is displayed for better visualization. (d) Color represents motion direction.  (e) Color represents visual word identity of each dense SIFT feature.  Temporal edge potential $\psi^t_{mn}$ depends on correspondence ratio on long trajectory and color affinity.  (f) Superpixels for corresponding vertices in the frame $\textit{f}$ are illustrated by object contours. For visualization purpose, it shows coarse grained superpixels.    Best viewed in color.}
		\label{fig:overview}
	\end{figure*}
	
	Video segmentation is one of the important problems in video understanding.  A video may contain a set of objects, from stationary to those undergoing dependent or independent motion.  Human understands a video by recognizing objects and infers the video context(i.e. what is happening in the video) by observing their motion.  Depending on the video context, parts or whole objects will have structured motion correlation.  However, there may be unrelated entities such as background or auxiliary objects which form additional structures as well.  Holistic representation of a video cannot effectively decompose and extract meaningful structure and it may increase intra-variability of a video class.  The goal of video segmentation is to obtain coherent object regions over frames so that a video can be represented as a set of objects and a meaningful structure can be extracted.  
	
	Ideally, the ultimate goal of video segmentation is to obtain pixelwise semantic segmentation of videos, where the objective is not only to partition a video into object regions but to infer object label of each region.  
	Semantic segmentation is actively investigated in urban driving scene understanding~\cite{Brostow08,Badrinarayanan10,Zhang10,Froehlich13}.  However, the urban scene videos contain rigid objects such as buildings, cars or road with typically smooth motion.
	In general, it is more challenging to segment and classify object regions in general, unconstrained videos.  First, the labor cost of obtaining pixelwise label annotation in video can be extremely high.  Instead, most of the datasets provide bounding box annotations on major objects without providing full frame coverage.  In addition, typical video datasets display high intra-class variability.  Objects or human subjects are deformable and their appearance would change due to changing illumination over frames.  Furthermore, motion pattern of objects in the same class of a video may exhibit idiosyncrasy.  Because of these aspects, learning a robust classifier for each and every object in a video remains, at present, an insurmountable task.
	
	Another fundamental challenge in video segmentation is that the inherent video object hierarchy may be highly subjective.  Annotations of multiple human annotators may vary significantly. For example, one annotator may assign a single label to the whole human body, whereas another annotator will label torso and leg part separately.  Furthermore, some objects may not have strong correlation to one feature alone.  For example, an object may have parts that show different color patterns but move consistently.  Hence, in practice, one may induce a hierarchical video segmentation with different levels of granularity from aggregated information of multi-cue feature channels. 
	
	In this paper, we propose a novel hierarchical video segmentation model which integrates temporal smooth labels with global structure consistency with preserving object boundaries.  Our contributions are as follows:
	\begin{itemize}
		\item We propose a video segmentation model that preserves multi-cue structures of object boundary and temporal smooth label with global spatio-temporal consistency.
		\item We propose an effective pairwise potential to represent spatio-temporal structure evaluated on object boundary, color, optical flow, texture and long trajectory correspondence.
		\item Video hierarchy is inferred through the process of graph edge consistency, which generalizes traditional hierarchy induction approaches.
		\item The proposed method infers precise coarse grained segmentation, where a segment may represent one whole object.
	\end{itemize}
	
	The remainder of this paper is organized as follows.  Section \ref{sec:rel_work} describes a set of related work and their limitations.  Our proposed model is introduced in Section \ref{sec:model}.  Experiments set up and results are described in Section \ref{sec:experiments}, followed by concluding remarks in Section \ref{sec:conclusion}.
	
	
	\section{Related Work}
	\label{sec:rel_work}
	One of the main objectives of video segmentation is to obtain spatio-temporal smoothness of the region labels.  Grundmann \etal~\cite{Grundmann10} proposed a greedy agglomerative clustering algorithm that merges two adjacent superpixels if their color difference is smaller than internal variance of each superpixel.  Granularity of the segmentation is controlled by adding a parameter to internal variance.  The algorithm obtains spatio-temporal smoothness on segment labels since it merges two adjacent superpixels.  In addition, it effectively detects newly appeared object due to the agglomerative clustering.  However, they only focus on color information without spatio-temporal structure.  As a consequence, it may merge a part of an object with another object or with the background, especially in the coarse-grained segmentation.  Furthermore, the approach does not extract object boundaries effectively because the algorithm does not make use of spatial structure from image gradients or edge detectors.
	
	Object boundary contour extracts spatial structure for image data.  Arbelaez \etal~\cite{Arbelaez11} introduced a hierarchical contour detector for image segmentation.  Their framework starts with best angular edge response on each pixel and agglomerative clustering constructs a hierarchical object contour map.  It is capable of detecting object boundaries even in a low contrast image where the object appearance is less distinctive to the background.  
	
	Moreover, the contour strength provides a cue to understand spatial structure.  It is likely that a strong contour separates an object to other objects, while a weak contour separates two parts inside of an object.  However, the algorithm is applicable only to image data and it is not trivial to extend to a video dataset.  The algorithm processes each video frame independently and produces object regions within each image.  It requires to match regions correspond to an object across frames to obtain temporal smoothness of segmentation.
	
	Galasso \etal~\cite{Galasso13} aim to obtain correspondence of superpixels across video frames by propagating labels from a source frame along the optical flow.  However, the quality of propagated labels typically decays due to flow estimation errors as the distance from the source frame increases.  They propose a remedy by propagating from the center frame, not taking into account global label consistency over the full video sequence.  Another limitation is that this label propagation approach cannot introduce objects because the label set of source frame does not contain a label corresponds to the new object.  In motion segmentation, Elqursh and Elgammal~\cite{Elqursh13} resolve the issue by splitting a group of trajectories if their dissimilarity becomes dominant.  However, the robustness of this approach depends highly on the choice of a threshold parameter, which needs to be tuned for each video.
	
	On the other hand, robust temporal structure information can be extracted from long-term trajectories.  Ochs \etal~\cite{Ochs14} introduce a video segmentation framework that depends on long-term point trajectories from large displacement optical flow~\cite{Brox11}.  They start with spatially sparse trajectory labels which are obtained by regularized spectral clustering on motion difference among trajectories.  Dense region labels are inferred by Potts energy minimization.  Although the proposed approach attains robust temporal consistency, it cannot distinguish objects of identical motion pattern because the trajectory label only depends on motion.  Nonetheless, the long trajectories offer a good cue to inferring long range temporal structure in a video.  For instance, two superpixels in distant frames can be hypothesized to have common identity if they share sufficiently many pixel trajectories.
	
	Galasso	\etal~\cite{Galasso12} aggregate a set of pairwise affinities in color, optical flow direction, long trajectory correspondence and adjacent object boundary.  With aggregated pairwise affinity, they adopt spectral clustering to infer segment labels.  Spectral clustering is one of the standard algorithms in the segmentation problem.  However, Nadler and Galun~\cite{Nadler07} illustrate cases where spectral clustering fails when the dataset contains structures at different scales of size and density for different clusters.  
	
	We propose a Markov Random Field(MRF) model whose vertices are defined from object contour based superpixel.  The model takes temporal smooth label likelihood as node potentials and global spatio-temporal structure information is incorporated as edge potentials in multi-modal feature channels, such as color, motion, object boundary, texture and long trajectories.  
	Since the proposed model takes contour based superpixels as vertices, the inferred segmentation preserves good object boundaries.  In addition, the model enhances long range temporal consistency over label propagation by incorporating global structure.  Moreover, we aggregate multi-modal features in the video so that the model can distinguish objects of identical motion.  Finally, MRF inference with unary and pairwise potential results in accurate segmentation compared to spectral clustering which only relies on pairwise relationship.
	
	As a result, the proposed model infers video segmentation labels by preserving accurate object boundaries which are locally smooth and consistent to global spatio-temporal structure of the video.
	
	
	\section{Proposed Model}
	\label{sec:model}
	\subsection{Multi-Cue Structure Preserving MRF Model}
	\label{sec:MSP-MRF}
	An overview of our framework for video segmentation is depicted in Figure \ref{fig:overview}.  A video is represented as a graph $\mathcal{G=(V,E)}$, where a vertex set $\mathcal{V}=\{ \mathcal{V}^1, \cdots,\mathcal{V}^F \}$ is defined on contour based superpixels from all frames $f \in \{1, \cdots, F\}$ in the video.  For each frame, an object contour map is obtained from contour detector~\cite{Arbelaez11}.  A region enclosed by a contour forms a superpixel.  An edge set $\mathcal{E} = \{ \mathcal{E}^s, \mathcal{E}^t \}$ describes relationship for each pair of vertices.  The edge set consists of spatial edges $ e_{ij} \in \mathcal{E}^s$ where $i,j \in \mathcal{V}^f$ and temporal edges $ e_{ij} \in \mathcal{E}^t$ where $i \in \mathcal{V}^f, j \in \mathcal{V}^{f'}, f \ne f'$.  
	
	Video segmentation is obtained by MAP inference on a Markov Random Field ${\bf Y} = \{y_i| i \in \mathcal{V}, y_i \in \mathcal{L} \}$ on this graph $\mathcal{G}$, where $P(Y) = \frac{1}{Z} \exp(-E(Y))$ and $Z$ is the partition function.  Vertex $i$ is labeled as $y_i$ from the label set $\mathcal{L}$ of size $L$.  MAP inference is equivalent to the following energy minimization problem.
	\begin{alignat}{3}
		\label{eq:model}
		\min ~ & E(Y) = \sum_{i \in \mathcal{V}} \boldsymbol{\phi}_i \cdot \mathbf{p}_i  + \sum_{(i,j)\in \mathcal{E}} &&\boldsymbol{\psi}_{ij}:\mathbf{q}_{ij},\\	
		\text{s.t.} & \sum_{l \in \mathcal{L}}p_i(l) = 1, && \forall i \in \mathcal{V}\\
		&\sum_{l' \in \mathcal{L}}q_{ij}(l,l')=p_i(l), \quad && \forall (i,j) \in \mathcal{E}, l \in \mathcal{L}\\
		&\mathbf{p}_i \in \{0,1\}^L, && \forall i \in \mathcal{V}\\
		&\mathbf{q}_{ij} \in \{0,1\}^{L \times L}, && \forall (i,j) \in \mathcal{E}
	\end{alignat}
	
	\noindent In (\ref{eq:model}), $\boldsymbol{\phi}_i$ represents node potentials for a vertex $i \in \mathcal{V}$ and $\boldsymbol{\psi}_{ij}$ is edge potentials for an edge $e_{ij} \in \mathcal{E}$.  As with the edge set $\mathcal{E}$, edge potentials are decomposed into spatial and temporal edge potentials, $\boldsymbol{\psi} = \{\boldsymbol{\psi}^s, \boldsymbol{\psi}^t\}$.  The vector $\mathbf{p}_i$ indicates label $y_i$ and $\mathbf{q}_{ij}$ is the label pair indicator matrix for $y_i$ and $y_j$.  Operators $\cdot$ and $:$ represent inner product and Frobenius product, respectively.  Spatial edge potentials are defined for each edge which connects the vertices in the same frame $i,j \in \mathcal{V}^f$.  In contrast, temporal edge potentials are defined for each pair of vertices in the different frames $i \in \mathcal{V}^f, j \in \mathcal{V}^{f'}, f \ne f'$.  It is worth noting that the proposed model includes spatial edges between two vertices that are not spatially adjacent and, similarly, temporal edges are not limited to consecutive frames.
	
	A set of vertices of the graph is defined from contour based superpixels such that the inferred region labels will preserve accurate object boundaries.  Node potential parameters are obtained from temporally smooth label likelihood.  Edge potential parameters aggregate appearance and motion features to represent global spatio-temporal structure of the video.  MAP inference of the proposed Markov Random Field(MRF) model will infer the region labels which preserve object boundary, attain temporal smoothness and are consistent to global structure.  Details are described in the following sections.
	
	
	\subsection{Node Potentials}
	\label{sec:np}
	Unary potential parameters $\boldsymbol{\phi}_i \in \mathbb{R}^{L}$ represent a cost of labeling vertex $i \in \mathcal{V}$ from a label set $\mathcal{L}$. While edge potentials represent global spatio-temporal structure in a video, node potentials in the proposed model strengthen temporal smoothness for label inference.  Temporal smooth label set $\mathcal{L}$ is obtained from a greedy agglomerative clustering~\cite{Grundmann10}.  The clustering algorithm merges two adjacent blobs in a video when color difference is smaller than the variance of each blob.  Node potential parameters $\boldsymbol{\phi}_i$ represent labeling cost of vertex $i$ from negative label likelihood $\mathbf{h}^l_i$.  
	\begin{align}
		\label{eq:node}
		\boldsymbol{\phi}_i &= - \mathbf{h}^l_i,\\
		\mathbf{h}^l_i &= [h^l_i(1), \cdots, h^l_i(L)]/H, \\
		H &= \sum_{b=1}^L h^l_i(b).
		\label{eq:indicator}
	\end{align}
	\noindent Each superpixel is evaluated by pixelwise cluster labels from $\mathcal{L}$ and the label histogram $\mathbf{h}_i$ represents label likelihood for the vertex $i$.
	As illustrated in Figure \ref{fig:overview}~(a), a superpixel has a mixture of pixelwise temporal smooth labels because the agglomerative clustering~\cite{Grundmann10} merges unstructured blobs.
	Let $h^l_i(b)$ be the number of pixelwise temporal smooth label $b$ in the corresponding superpixel of vertex $i$.
	As described in \ref{sec:MSP-MRF}, a vertex is defined on a superpixel which is enclosed by an object contour.  Arbelaez \etal~\cite{Arbelaez11} extract object contours so that taking different threshold values on the contours will produce different granularity levels of enclosed regions.  In our proposed model, we take a set of vertices $\mathcal{V}^f$ from a video frame $f$ by a single threshold on contours which results in fine-grained superpixels.  
	
	
	\subsection{Spatial Edge Potentials}
	Binary edge potential parameters $\psi$ consist of two different types; spatial and temporal edge potentials, $\psi^s$ and $\psi^t$, respectively .  Spatial edge potentials $\psi_{ij}^s$ model pairwise relationship of two vertices $i$ and $j$ within a single video frame $f$.  We define these pairwise potentials as follows:
	\begin{align}
		\label{eq:spat_edge}
		\boldsymbol{\psi}_{ij}^s(l,l') &= \left\{ 
		\begin{array}{ l l}
			\frac{ \psi_{ij}^b+\psi_{ij}^c+\psi_{ij}^o+\psi_{ij}^x}{4} & \quad \text{if $l \ne l',~ \psi_{ij}^s \ge \tau$},\\
			0 & \quad \text{otherwise}
		\end{array} \right.
	\end{align}
	\noindent A spatial edge potential parameter $\boldsymbol{\psi}_{ij}^s (l,l')$ is the $(l,l')$ element of $\mathbb{R}^{L \times L}$ matrix which represents the cost of labeling a pair of vertices $i$ and $j$ as $l$ and $l'$, respectively.   It takes Potts energy where all different pairs of label take homogeneous cost $\boldsymbol{\psi}_{ij}^s$.  Spatial edge potentials $\boldsymbol{\psi}^s$ are decomposed into $\psi^b, \psi^c, \psi^o, \psi^x$, which represent pairwise potentials in the channel of object boundary, color, optical flow direction and texture.  Pairwise cost of having different labels is high if the two vertices $i$ and $j$ have high affinity in the corresponding channel.  As a result, edge potentials increase the likelihood of assigning the same label to vertices $i$ and $j$ during energy minimization. 
	
	The edge potentials take equal weights on all channels.  Importance of each channel may depend on video context and different videos have dissimilar contexts.  Learning weights of each channel is challenging and it is prone to overfitting due to high variability of video context and limited number of labeled video samples in the dataset.  Hence, the propose model equally weights all channels.
	
	The model controls the granularity of segmentation by a threshold $\tau$.  In (\ref{eq:spat_edge}), the pairwise potential is thresholded by $\tau$.  If $\tau$ is set to a high value, only edges with higher affinity will be included in the graph.  On the other hand, if we set a low value to $\tau$, the number of edges increases and more vertices will be assigned to the same label because they are densely connected by the edge set.   We next discuss each individual potential type in the context of our video segmentation model.
	
	\noindent{\bf Object Boundary Potentials $\boldsymbol{\psi}^b$.}
	Object boundary potentials $\boldsymbol{\psi}^b_{ij}$ evaluate cost of two vertices $i$ and $j$ in the same frame assigned to different labels in terms of object boundary information.  The potential parameters are defined as follows:
	\begin{align}
		\boldsymbol{\psi}_{ij}^b &= \exp(- d_{\text{MMPW}}(i,j)/\gamma_b).
	\end{align}
	\noindent where $d_{\text{MMPW}}(i,j)$  represents the minimum boundary path weight among all possible paths from a vertex $i$ to $j$.  The potentials $\psi^b$ are obtained from Gaussian Radial Basis Function(RBF) of $d_{\text{MMPW}}(i,j)$ with $\gamma_b$ which is the mean of $d_{\text{MMPW}}(i,j)$ as a normalization term.
	
	If the two superpixels $i$ and $j$ are adjacent, their object boundary potentials are decided by the shared object contour strength $b({e_{ij})}$, where $e_{ij}$ is the edge connects vertices $i$ and $j$ and the boundary strength is estimated from contour detector~\cite{Arbelaez11}.  The boundary potentials can be extended to non-adjacent vertices $i$ and $j$ by evaluating a path weight from vertex $i$ to $j$.  For each path $p$ from a vertex $i$ to $j$, boundary potential of path $p$ is evaluated by taking the maximum edge weights $b(e_{uv})$ where $e_{uv}$ is an edge along the path $p$.   The algorithm to calculate $d_{\text{MMPW}}(i,j)$ is described in Algorithm \ref{algo:contour_path}, which modifies Floyd-Warshall shortest path algorithm.
	
	Typically, a path in a graph is evaluated by sum of edge weights along the path.  However, in case of boundary strength between the two non-adjacent vertices in the graph, total sum of the edge weights along the path is not an effective measurement because the sum of weights is biased toward the number of edges in the path.  For example, a path consists edges of weak contour strength may have the higher path weight than another path which consists of smaller number of edges with strong contour.
	Therefore, we evaluate a path by the maximum edge weight along the path and the path weight is govern by an edge of the strongest contour strength.
	
	\setlength{\textfloatsep}{12pt}
	\begin{algorithm}[t]
		\caption{Minimum Max-edge Path Weight}\label{euclid}
		\label{algo:contour_path}	
		\begin{algorithmic}[1]
			\Procedure{MMPW}{$\mathcal{V},\mathcal{E}$}
			\State $d\gets \infty$
			\For {$v \in \mathcal{V} $}
			\State $d[v][v]\gets 0$
			\EndFor\label{euclidendwhile}
			\For {$(u,v) \in \mathcal{E} $}
			\State $d[u][v]\gets b({e_{uv}})$ \Comment{assign boundary score}
			\EndFor\label{euclidendwhile}		
			\For {$k \in \mathcal{V} $}
			\For {$i \in \mathcal{V} $}
			\For {$j \in \mathcal{V} $}
			\If{$d[i][j] > \max(d[i][k], d[k][j])$}
			\State $d[i][j] \gets \max(d[i][k], d[k][j])$
			\EndIf
			\EndFor
			\EndFor
			\EndFor
			\State \textbf{return} $d_\text{MMPW} \gets d$\
			\EndProcedure
		\end{algorithmic}
	\end{algorithm}
	
\begin{figure*}[ht]
		\centering
		\begin{subfigure}{0.24\textwidth}
			\includegraphics[width=\textwidth]{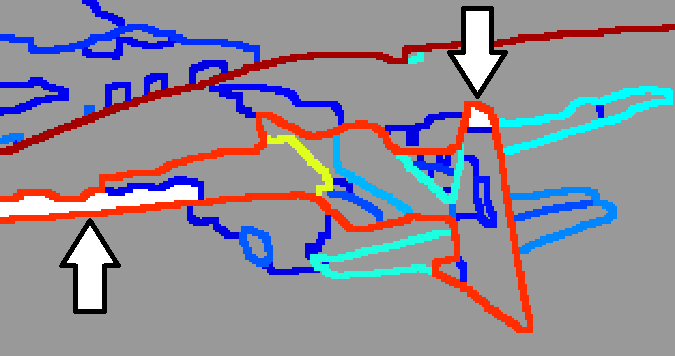}
			\caption{Contour strength}
		\end{subfigure}
		\begin{subfigure}{0.24\textwidth}
			\includegraphics[width=\textwidth]{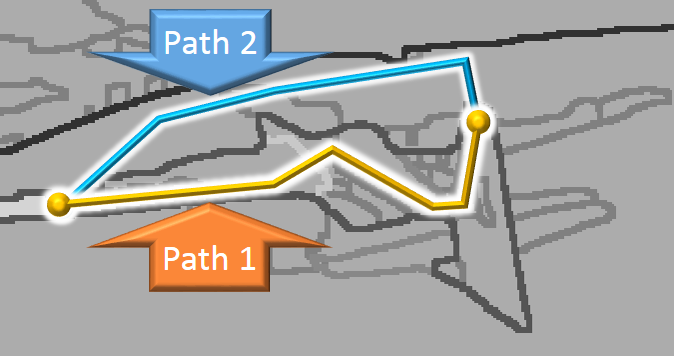}
			\caption{Two contour paths}
		\end{subfigure}
		\begin{subfigure}{0.24\textwidth}
			\includegraphics[width=\textwidth]{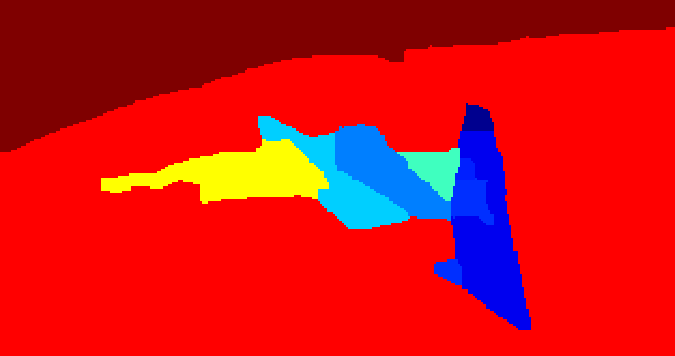}
			\caption{Edge sum path weight}
		\end{subfigure}
		\begin{subfigure}{0.24\textwidth}
			\includegraphics[width=\textwidth]{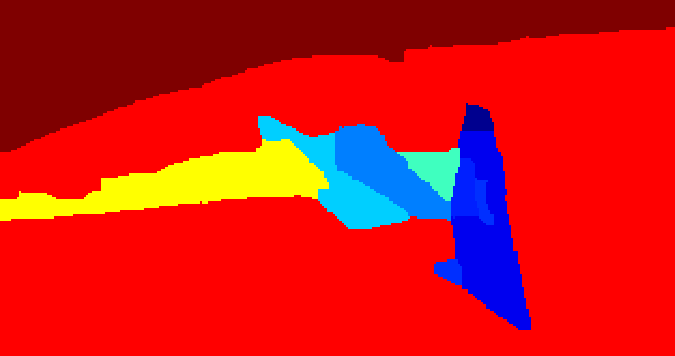}
			\caption{Max edge path weight}
		\end{subfigure}
		\caption{Comparison of two types of path weight models. }
		\vspace{-0.5em}
		\label{fig:maxWgtPath}
		
	\end{figure*}

	Figure \ref{fig:maxWgtPath} illustrates two different path weight models of the max edge weight and the sum edge weight.   Figure \ref{fig:maxWgtPath} (a) illustrates contour strength where red color represents high strength.  Two vertices indicated by white arrows are selected in an airplane.  In Figure \ref{fig:maxWgtPath} (b), two paths are displayed.  \textit{Path 2} consists of less number of edges but it intersects with a strong contour that represents boundary of the airplane.  If we evaluate object boundary score between the two vertices, \textit{Path 1} should be considered since it connects vertices within the airplane.  Figure \ref{fig:maxWgtPath} (c) shows edge sum path weight from a vertex at tail to all the other vertices.  It displays that the minimum path weight between the two vertices are evaluated on \textit{Path 2}.  On the other hand, Figure \ref{fig:maxWgtPath} (d) illustrates that max edge path weight takes \textit{Path 1} as minimum path weight which conveys human perception of object hierarchy.
	
	\noindent{\bf Color Potentials $\boldsymbol{\psi}^c$.}
	Color feature for each vertex is represented by a histogram of CIELab color space in the corresponding superpixel.  Color potential $\psi^c_{ij}$ between the vertex $i$ and $j$ is evaluated on two color histograms $\mathbf{h}^c_i$ and $\mathbf{h}^c_j$:
	\begin{align}
		\boldsymbol{\psi}_{ij}^c &= \exp(-d_\text{EMD}(\mathbf{h}^c_i, \mathbf{h}^c_j)/\gamma_c).
	\end{align}
	\noindent where $d_\text{EMD}(\mathbf{h}^c_i, \mathbf{h}^c_j)$ is Earth Mover's Distance(EMD) between $\mathbf{h}^c_i$ and $\mathbf{h}^c_j$ of vertices $i$ and $j$ and $\gamma_c$ is the normalization parameter.
	
	Earth Mover's Distance~\cite{Pele09} is a distance measurement between two probability distributions.  EMD is typically more accurate over $\chi^2$ distance in color space of superpixels. An issue with $\chi^2$ distance is that if the two histograms on simplex do not share non-zero color bins, the two histogram are evaluated with the maximum distance of $1$.  Therefore, distance of vertices $i$ and $j$ is the same as the distance between $i$ and $k$, if $i,j,k$ do not share any color bins.  This occurs often when we compare color feature of superpixels because superpixel is intended to exhibit coherent color especially in the fine grained level.  Superpixels on different objects or different parts of an object may have different colors.  For example, if we use $\chi^2$ distance to measure color difference of superpixels, distance between superpixels of red and orange will have the same distance of red and blue because they do not share color bins.  However, this is not intuitive to human perception.  In contrast, EMD considers distance among each color bin, hence it is able to distinguish non overlapping color histograms.  
	
	
	\noindent{\bf Optical Flow Direction Potentials $\boldsymbol{\psi}^o$.}
	In each video frame, motion direction feature of $i$th vertex can be obtained from a histogram of optical flow direction $\mathbf{h}^o_i$.  As with the case of color potentials, we use EMD between the two histograms $\mathbf{i}^o_i$ and $\mathbf{i}^o_j$ to accurately estimate difference direction in motion:
	\begin{align}
		\boldsymbol{\psi}_{ij}^o &= \exp(-d_\text{EMD}(\mathbf{h}^o_i, \mathbf{h}^o_j)/\gamma_o)
	\end{align}	
	\noindent where $\gamma_o$ is the mean EMD distance on optical flow histogram.
	
	
	\noindent{\bf Texture Potentials $\boldsymbol{\psi}^x$.}
	Dense SIFT features are extracted for each superpixel and Bag-of-Words(BoW) model is obtained from K-means clustering on D-SIFT features.  We evaluate SIFT feature on multiple dictionaries of different $K$.  Texture potentials $\boldsymbol{\psi}^x$ are calculated from RBF on $\chi^2$ distance of two BoW histograms $\mathbf{h}^x_i$ and $\mathbf{h}^x_j$, which is a typical choice of distance measurement for BoW model:
	\begin{align}
		\boldsymbol{\psi}_{ij}^x &= \exp(-d_{\chi^2} (\mathbf{h}^x_i, \mathbf{h}^x_j) / \gamma_x )
	\end{align}	
	\noindent where parameter $\gamma_x$ is the mean $\chi^2$ distance on D-SIFT word histogram.
	
	
	\subsection{Temporal Edge Potentials}
	Temporal edge potentials define correspondence of vertices at different frames.  It relies on long trajectories which convey long range temporal dependencies and more robust than optical flow.
	\begin{align}
		\label{eq:temp_edge}
		\boldsymbol{\psi}_{ij}^t(l,l') &= \left\{ 
		\begin{array}{ l l}
			\psi_{ij}^t & \quad \text{if $l \ne l'$},\\
			0 & \quad \text{otherwise}
		\end{array} \right. , \\
		\psi_{ij}^t &= \left\{
		\begin{array}{l l}
			\frac{\psi_{ij}^r+\psi_{ij}^c}{2} & \quad \text{if $\psi_{ij}^t \ge \tau$} \\
			0 &  \quad \text{otherwise}
		\end{array} \right., \\
		\psi_{ij}^r &= \frac{|T_i \cap T_j|}{|T_i \cup T_j|},\\
		\psi_{ij}^c &= \exp(-d_\text{EMD}(\mathbf{h}^c_i, \mathbf{h}^c_j)/\gamma_c).
	\end{align}
	\noindent where $T_i$ is a set of long trajectories which pass through vertex $i$.  Pairwise potential $\psi_{ij}^r$ represents temporal correspondence of two vertices from overlapping ratio of long trajectories that vertices $i$ and $j$ shares, where $i \in \mathcal{V}^f, j \in \mathcal{V}^{f'}$ and $f \ne f'$. In order to distinguish two different objects of the same motion, we integrate color potentials $\psi^c$ between two vertices.  Long trajectories are extracted from ~\cite{Brox10}.
	
	
	\subsection{Hierarchical Inference on Segmentation Labels}
	\label{sec:hierarchi_infer}
	The proposed model attains hierarchical inference of segmentation labels by controlling the number of edges with a fixed set of vertices defined at a finest level of superpixels.  As the edge set becomes dense in the graph, the energy function in (\ref{eq:model}) takes higher penalties from the pairwise potentials.  As a consequence, vertices connected by dense edges will be assigned to the same label and it leads to coarse-grained segmentation.
	
	In contrast, another approach that enables hierarchical segmentation is to define a hierarchical vertex set in a graph.  A set of vertices in the finer level will be connected to a vertex in coarser level.  It introduces another set of edges which connect vertices at different levels of hierarchy.
	
	Our proposed approach on hierarchical inference takes computational advantages over graph representation with a hierarchical vertex set.  Our proposed graph representation has less the number of vertices and edges because we have a single finest level of hierarchy without additional vertices for coarser levels.  This advantage not only enables an efficient graph inference, but also take less computation time to calculate node and edge potentials for additional vertex and edge sets.
	
	
	\section{Experimental Evaluation}
	\label{sec:experiments}
	\subsection{Dataset} 
	\label{sec:dataset}
	We evaluate the proposed model on VSB100 video segmentation benchmark data provided by Galasso \etal~\cite{Galasso13}.  There are a few additional video datasets which have pixelwise annotation.  FBMS-59 dataset~\cite{Ochs14} consists of 59 video sequences and SegTrack v2 dataset~\cite{Li13} consists of 14 sequences.  However, the both datasets annotate on a few major objects leaving whole background area as one label.  It is more appropriate for object tracking or background subtraction task.  On the other hand, VSB100 consists of 60 test video sequences of maximum 121 frames.  For each video, every 20 frame is annotated with pixelwise segmentation labels by four annotators.  The dataset contains the largest number of video sequences annotated with pixelwise label, which allows quantitative analysis.  The dataset provides a set of evaluation measurements.
	
	\textbf{Volume Precision-Recall.}  VPR score measures overlap of the volume between the segmentation result of the proposed algorithm $\mathbb{S}$ and ground truths $\{\mathbb{G}_i\}_{i=1}^M$ annotated by $M$ annotators.  Over-segmentation will have high precision with low recall score.
	
	\textbf{Boundary Precision-Recall.}  BPR score measures overlap between object boundaries of the segmentation result $S$ and ground truths boundaries $\{G_i\}_{i=1}^M$.  Conversely to VPR, over-segmentation will have low precision with high recall scores.
	
	
	\subsection{MSP-MRF Setup} 
	In this section, we present the detailed setup of our Multi-Cue Structure Preserving Markov Random Field (MSP-MRF) model for unconstrained video segmentation problem.  
	As described in Section \ref{sec:np}, we take a single threshold on image contour, so that each frame contains approximately 100 superpixels.  We assume that this granularity level is fine enough such that no superpixel at this level will overlay on multiple ground truth regions.  Node potential (\ref{eq:node}) is evaluated for each superpixel with temporal smooth label obtained with agglomerative clustering~\cite{Grundmann10}.  Although we chose the $11$th fine grained level of hierarchy, Section \ref{sec:gl} illustrates that the proposed method shows stable performance over different label set size $|\mathcal{L}|$ for node potential.
	Finally, edge potential is estimated as in (\ref{eq:spat_edge}), (\ref{eq:temp_edge}).  For color histograms, we used 50 bins for each CIELab color channel.  In addition, 50 bins were set for horizontal and vertical motion of optical flow.  For D-SIFT Bag-of-Words model, we used 5 dictionaries of $K=100,200,400,800,1000$ words.  Energy minimization problem in (\ref{eq:model}) for MRF inference is optimized using FastPD algorithm~\cite{Komodakis07}.
	
	
	\begin{figure}
		\begin{center}
			\begin{tabular}{l @{}  m{2.2cm} m{2.2cm} m{2.2cm}}
				\parbox[c]{4mm}{\multirow{1}{*}{\rotatebox[origin=c]{90}{Video}}}&
				\includegraphics[width=2.5cm]{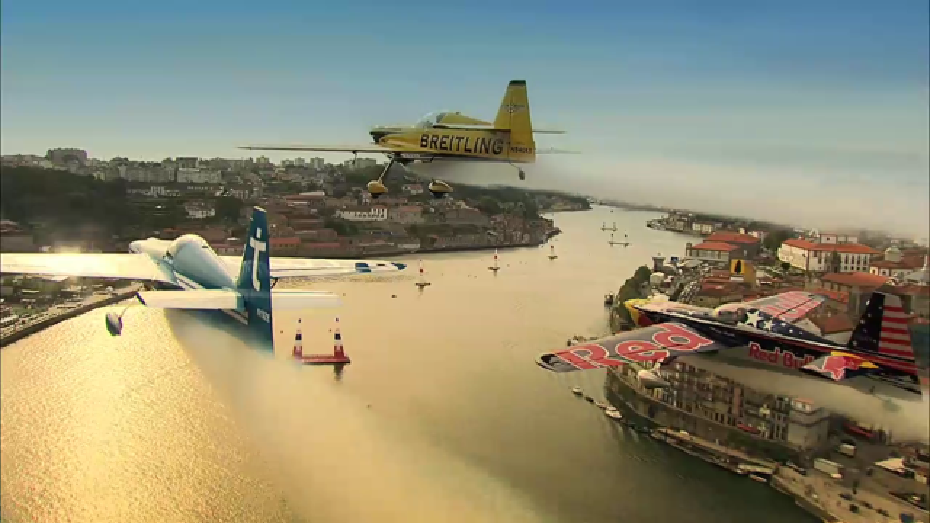} &
				\includegraphics[width=2.5cm]{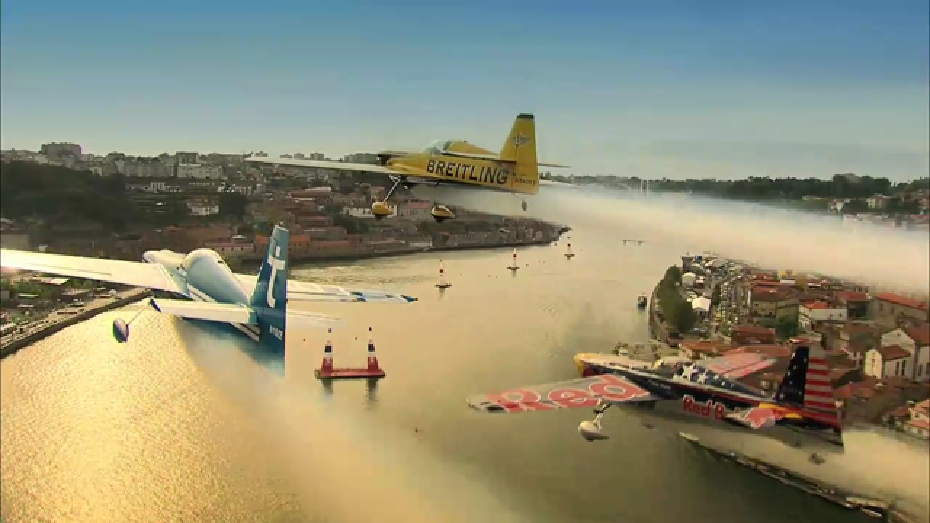} &
				\includegraphics[width=2.5cm]{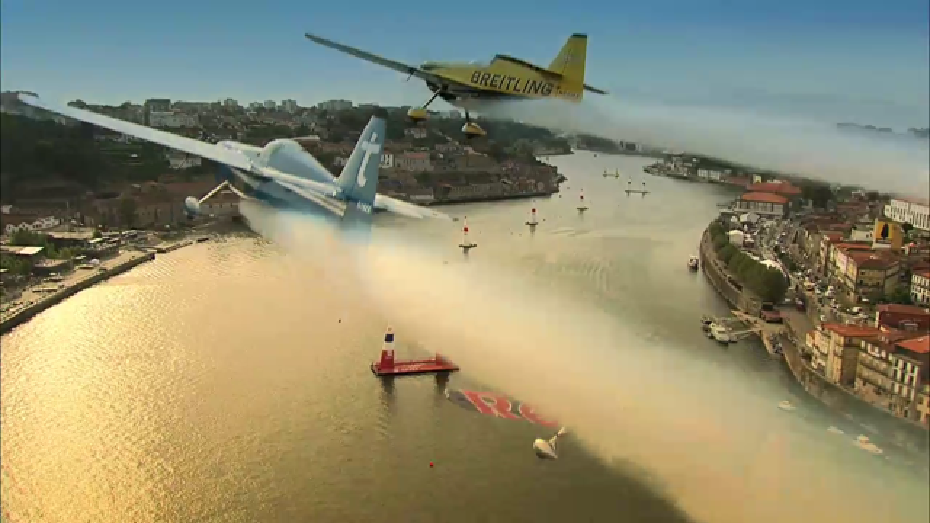} \\ 
				
				\parbox[c]{4mm}{\multirow{1}{*}{\rotatebox[origin=c]{90}{GT}}}&
				\includegraphics[width=2.5cm]{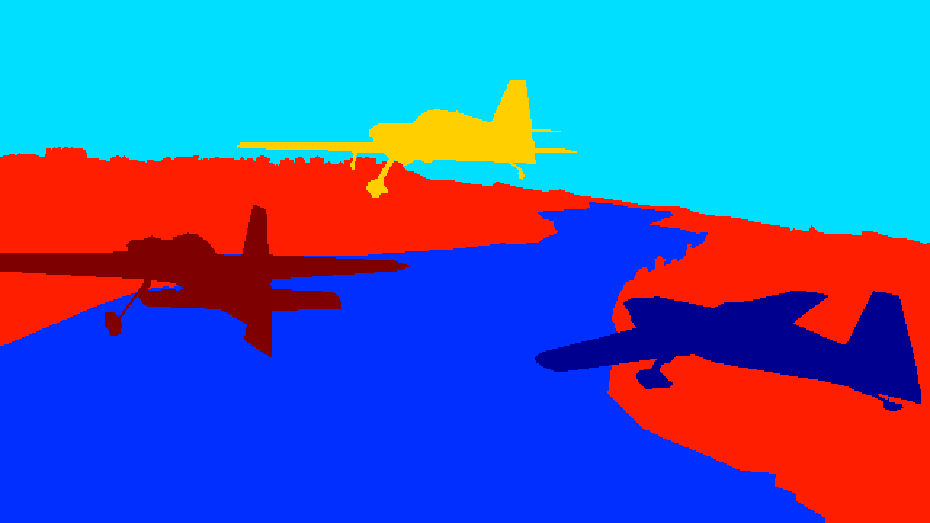} &
				\includegraphics[width=2.5cm]{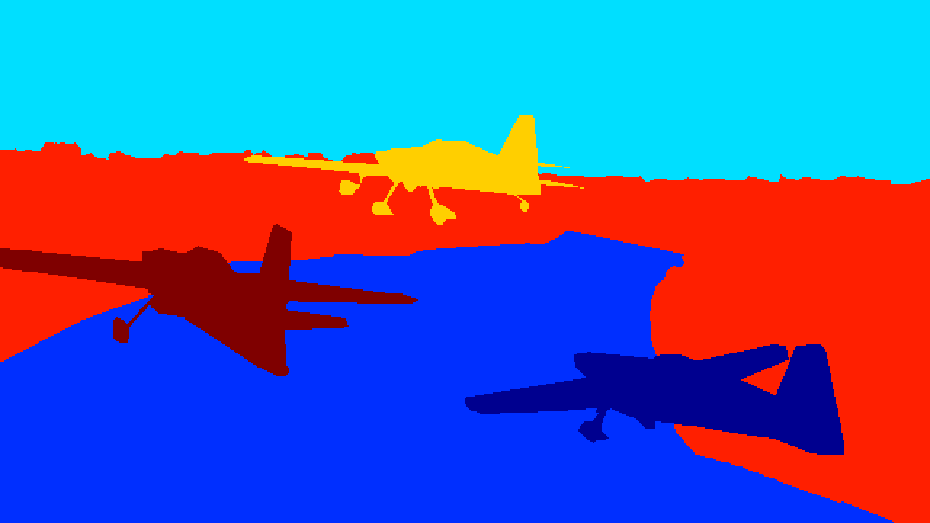} &
				\includegraphics[width=2.5cm]{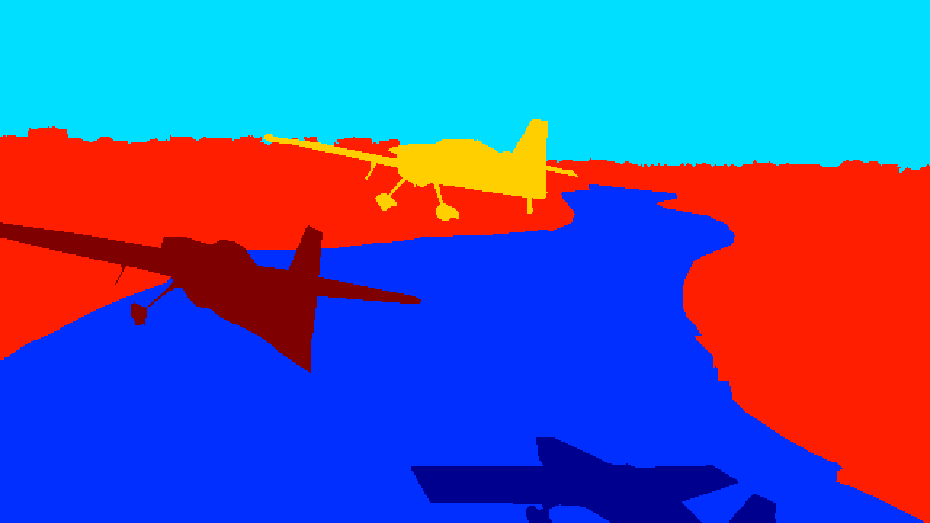} \\ 			
				
				\parbox[c]{4mm}{\multirow{1}{*}{\rotatebox[origin=c]{90}{\cite{Grundmann10}}}}&						
				\includegraphics[width=2.5cm]{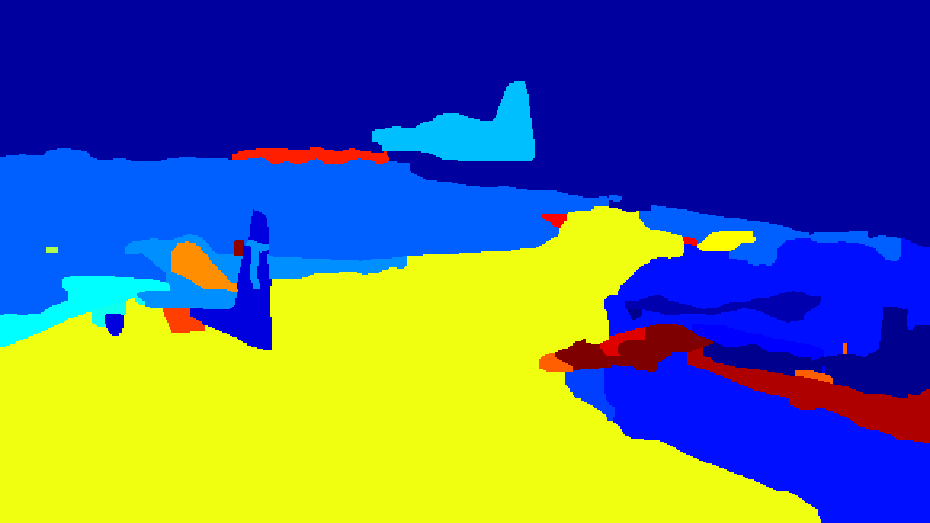} &
				\includegraphics[width=2.5cm]{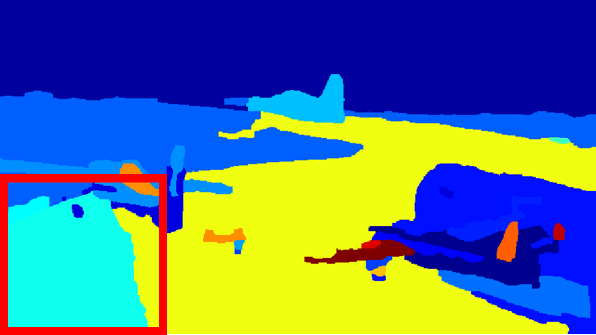} &
				\includegraphics[width=2.5cm]{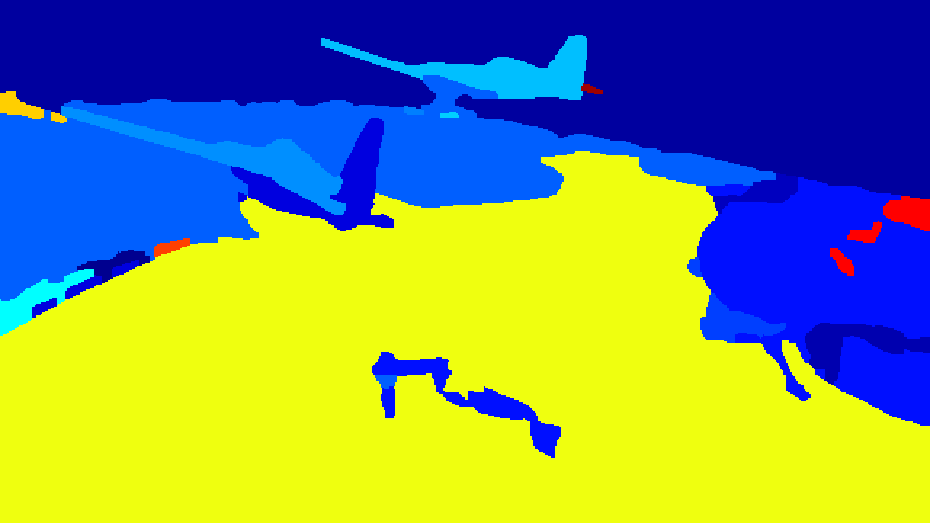} \\ 		
				
				\parbox[c]{4mm}{\multirow{1}{*}{\rotatebox[origin=c]{90}{Ours}}}&			
				\includegraphics[width=2.5cm]{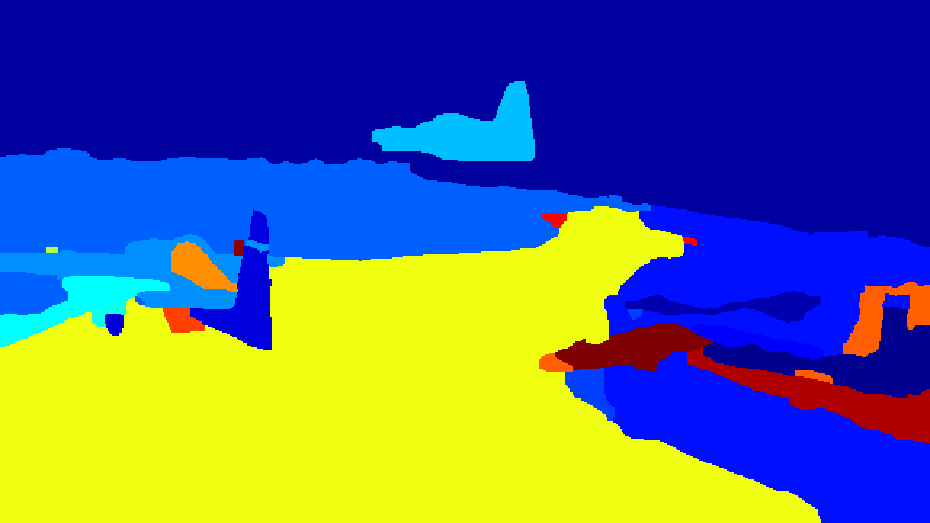} &
				\includegraphics[width=2.5cm]{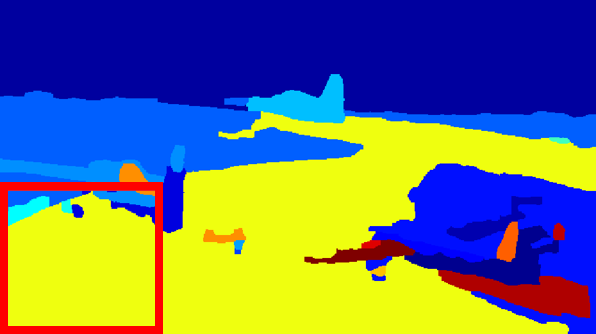} &
				\includegraphics[width=2.5cm]{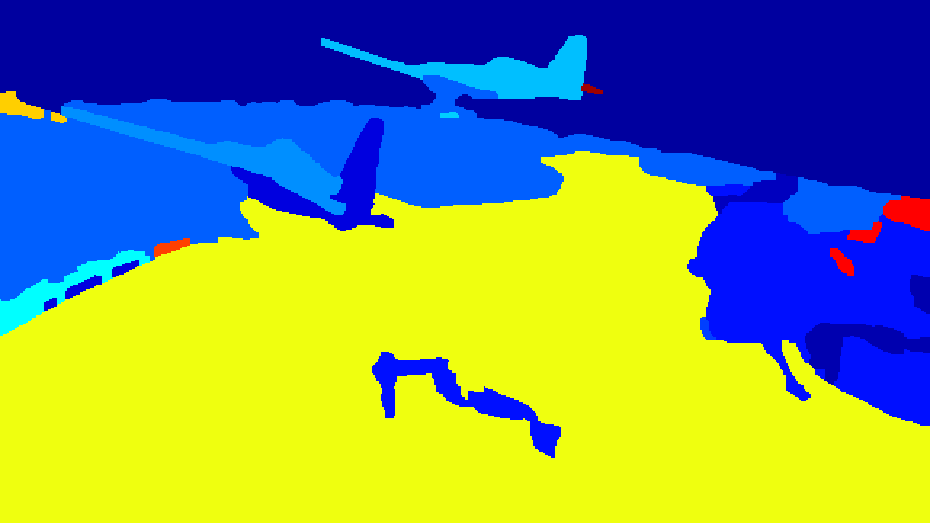} \\ 									
			\end{tabular}
			\caption{Temporal consistency recovered by MSP-MRF.}
			\label{fig:airplane}	
			\vspace{-1.5em}	
		\end{center}
	\end{figure}	
	
	\begin{figure}[!t]
		\begin{center}
			\vspace{-0.3em}
			\begin{tabular}{l @{}  m{2.2cm} m{2.2cm} m{2.2cm}}
				& \multicolumn{1}{c}{Video} & \multicolumn{1}{c}{Ground Truth} &\\
				&
				\includegraphics[width=2.5cm]{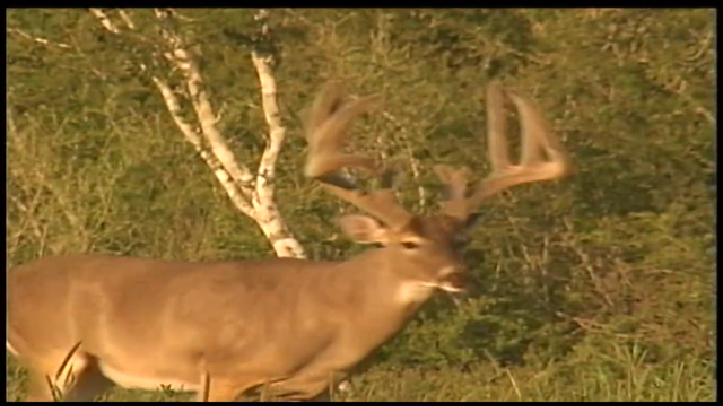} &
				\includegraphics[width=2.5cm]{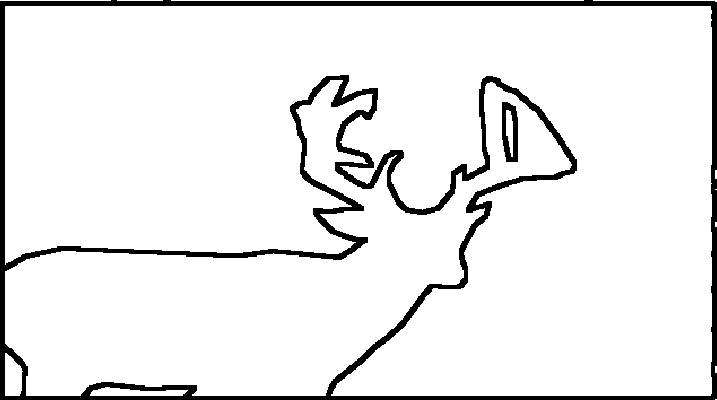} &\\ 
				& \multicolumn{1}{c}{Coarse} & \multicolumn{1}{c}{Mid} & \multicolumn{1}{c}{Fine} \\
				
				\parbox[c]{4mm}{\multirow{1}{*}{\rotatebox[origin=c]{90}{\cite{Grundmann10}}}}&					
				\includegraphics[width=2.5cm]{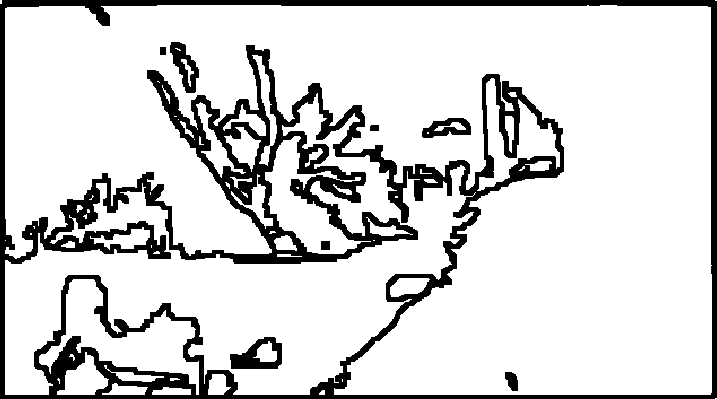} &
				\includegraphics[width=2.5cm]{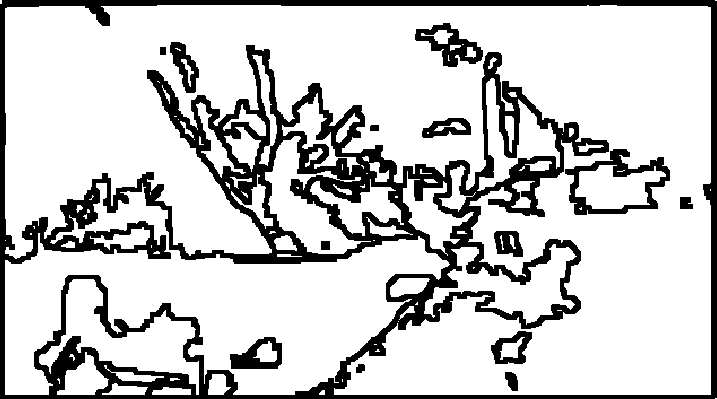} &
				\includegraphics[width=2.5cm]{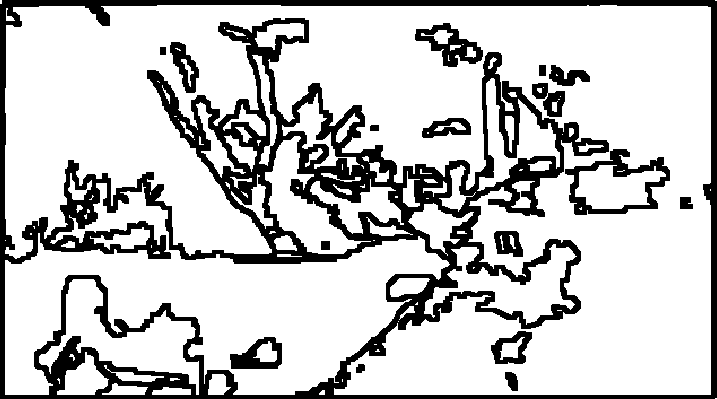} \\ 
				
				\parbox[c]{4mm}{\multirow{1}{*}{\rotatebox[origin=c]{90}{Ours}}}&			
				\includegraphics[width=2.5cm]{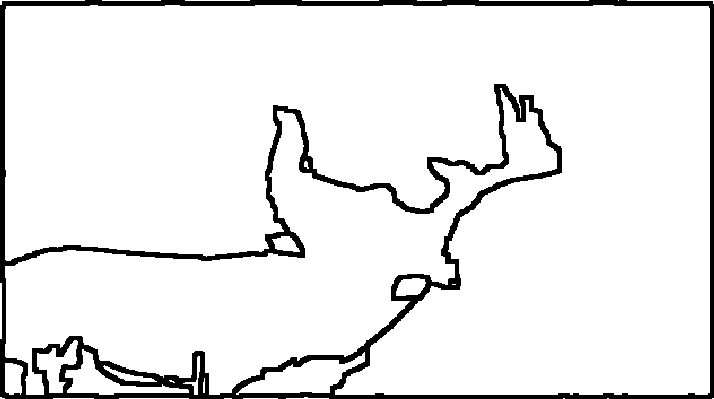} &
				\includegraphics[width=2.5cm]{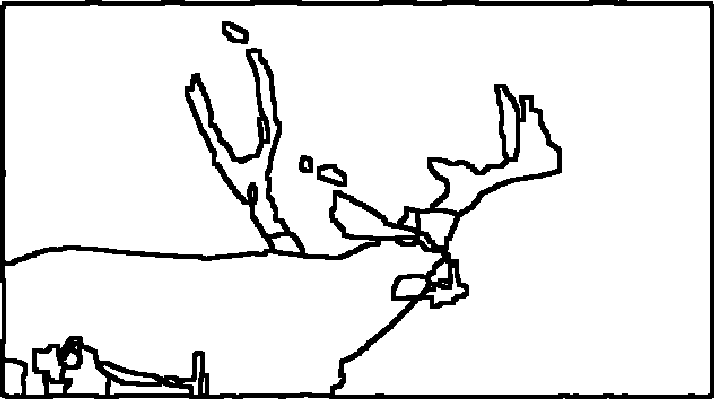} &
				\includegraphics[width=2.5cm]{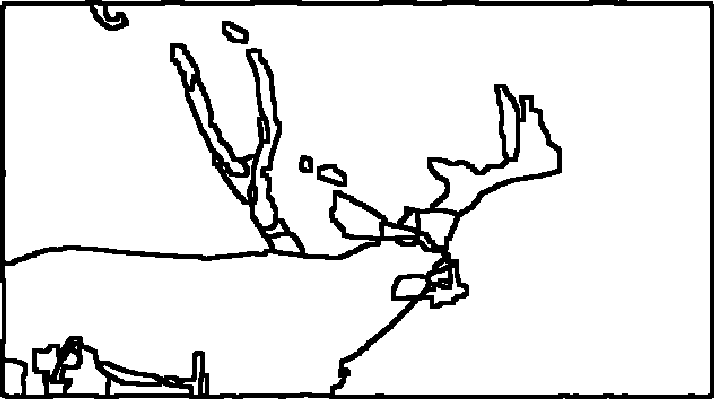} \\			
				\vspace{-0.7em}		\\	  \hline
				& \multicolumn{1}{c}{Video} & \multicolumn{1}{c}{Ground Truth} &\\
				&
				\includegraphics[width=2.5cm]{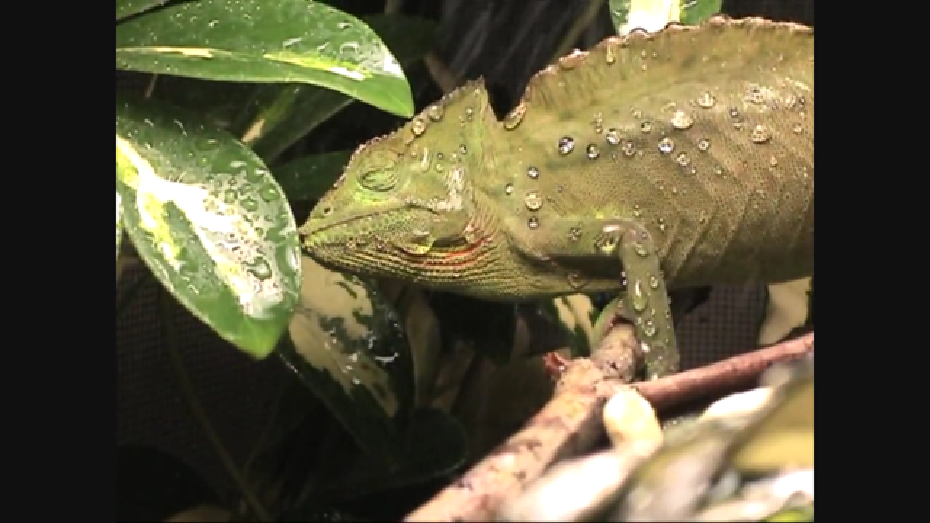} &
				\includegraphics[width=2.5cm]{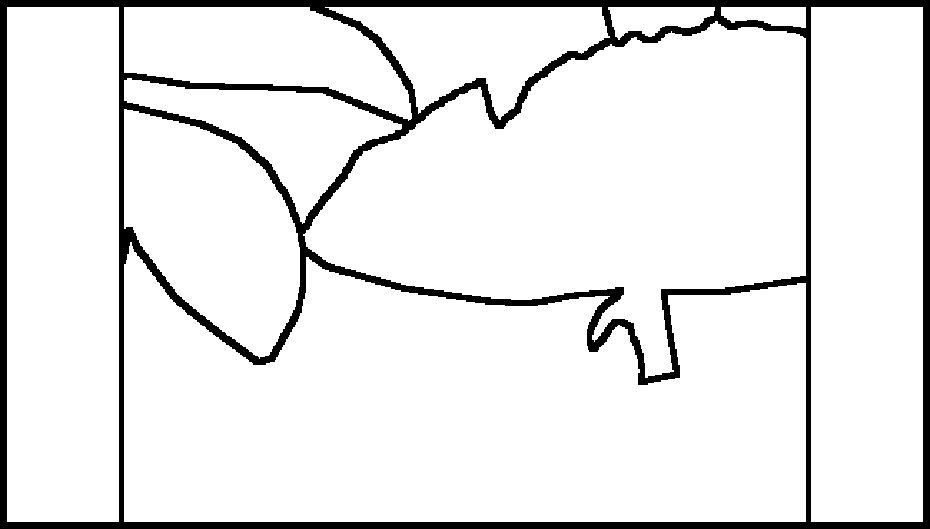} &\\ 
				& \multicolumn{1}{c}{Coarse} & \multicolumn{1}{c}{Mid} & \multicolumn{1}{c}{Fine} \\
				
				\parbox[c]{4mm}{\multirow{1}{*}{\rotatebox[origin=c]{90}{\cite{Grundmann10}}}}&					
				\includegraphics[width=2.5cm]{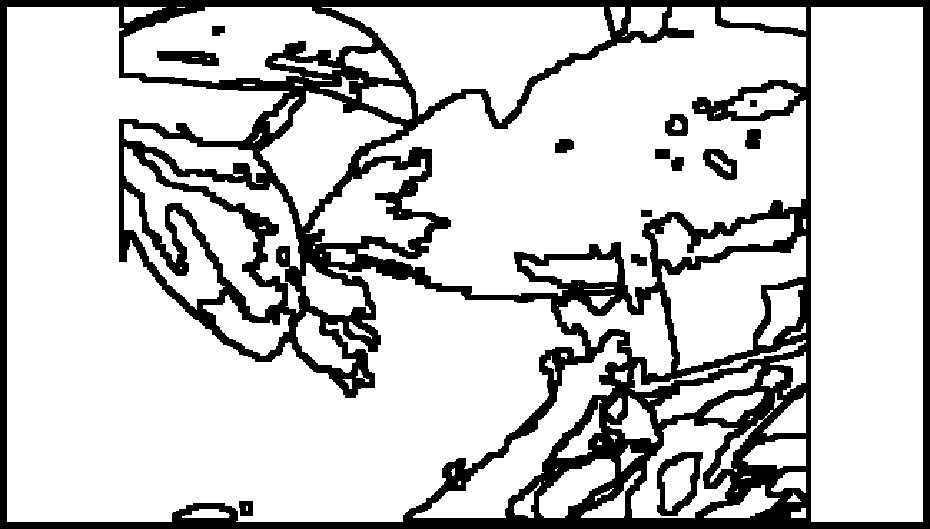} &
				\includegraphics[width=2.5cm]{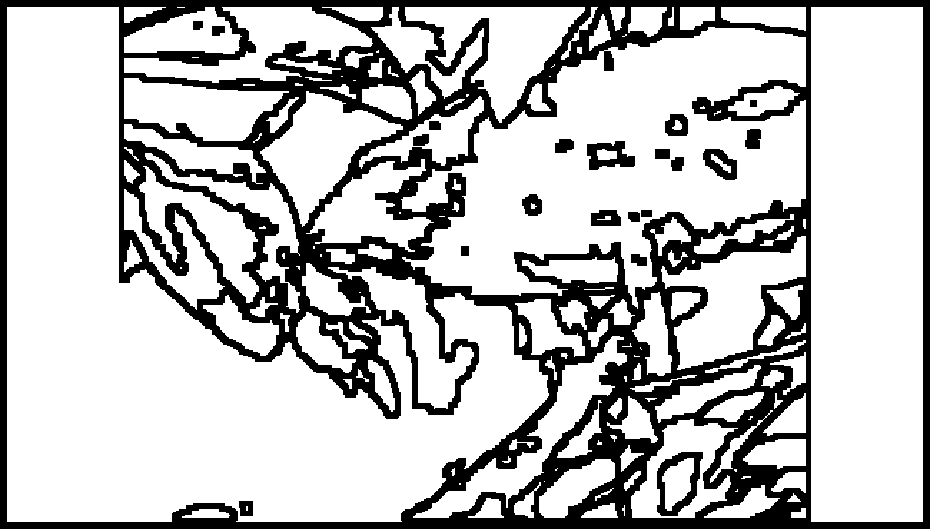} &
				\includegraphics[width=2.5cm]{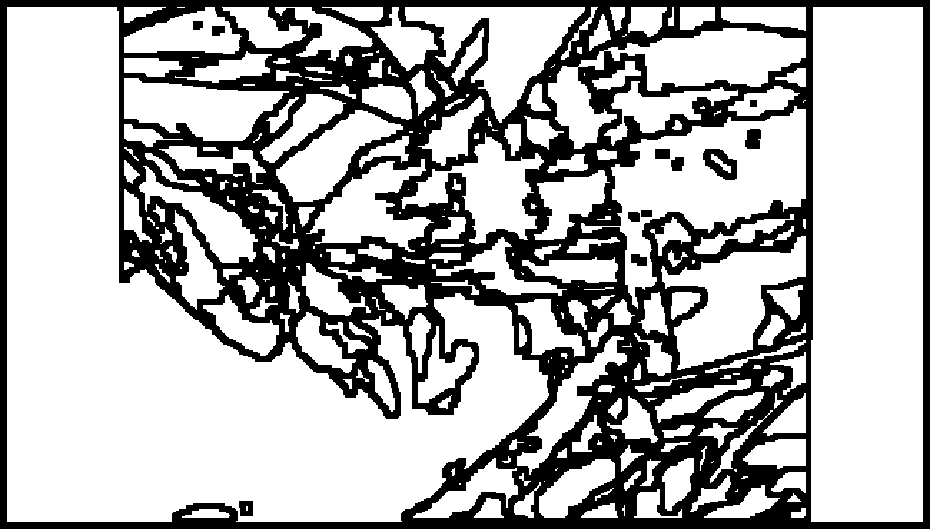} \\ 
				
				\parbox[c]{4mm}{\multirow{1}{*}{\rotatebox[origin=c]{90}{Ours}}}&			
				\includegraphics[width=2.5cm]{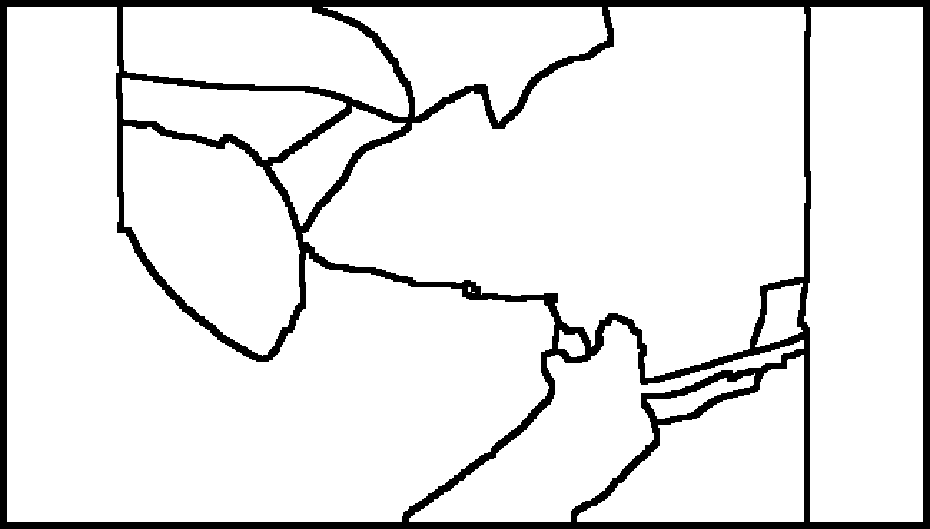} &
				\includegraphics[width=2.5cm]{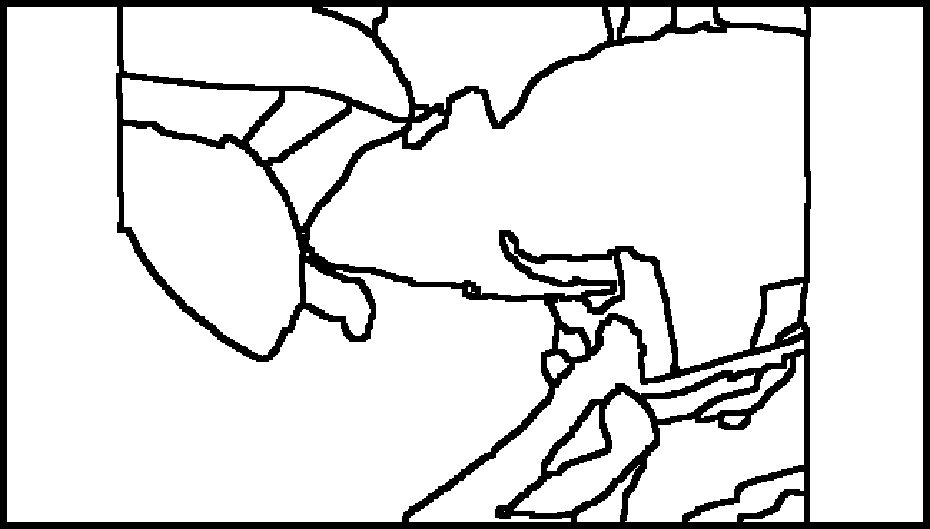} &
				\includegraphics[width=2.5cm]{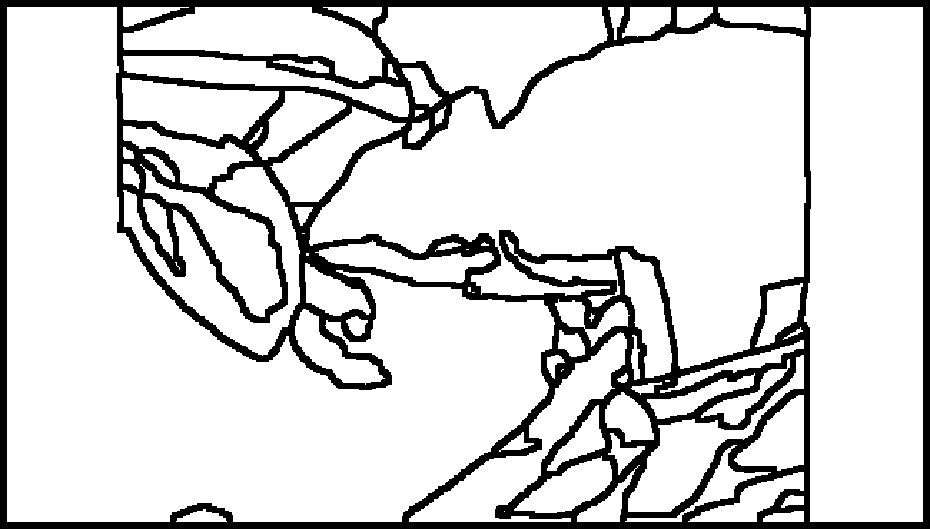} \\											
			\end{tabular}
			\vspace{-1em}
			\caption{Comparison of segmentation boundary on the same granularity levels on two videos.}
			\label{fig:res_vid1}		
			\vspace{-1em}
		\end{center}
	\end{figure}	
	
	
	\subsection{Qualitative Analysis}
	Figure \ref{fig:airplane} illustrates a segmentation result on an {\it airplane} video sequence.  MSP-MRF rectifies temporally inconsistent segmentation result of \cite{Grundmann10}.  For example, in the fourth column of Figure \ref{fig:airplane}, the red bounding boxes show MSP-MRF rectified label from Grundmann's result such that labels across frames become spatio-temporally consistent.
	
	In addition, control parameter $\tau$ successfully obtains different granularity level of segmentation.  For MSP-MRF, the number of region labels is decreased as $\tau$ decreases.  Figure \ref{fig:res_vid1} compares video segmentation results of MSP-MRF with Grundmann's by displaying segmentation boundary on the same granularity levels, where the two methods have the same number of segments in the video.  MSP-MRF infers spatial smooth object regions, which illustrates the fact that the proposed model successfully captures spatial structure of objects.
	
	\begin{figure}[!tbp]
		\centering
		\begin{subfigure}{0.242\textwidth}
			\includegraphics[width=\textwidth]{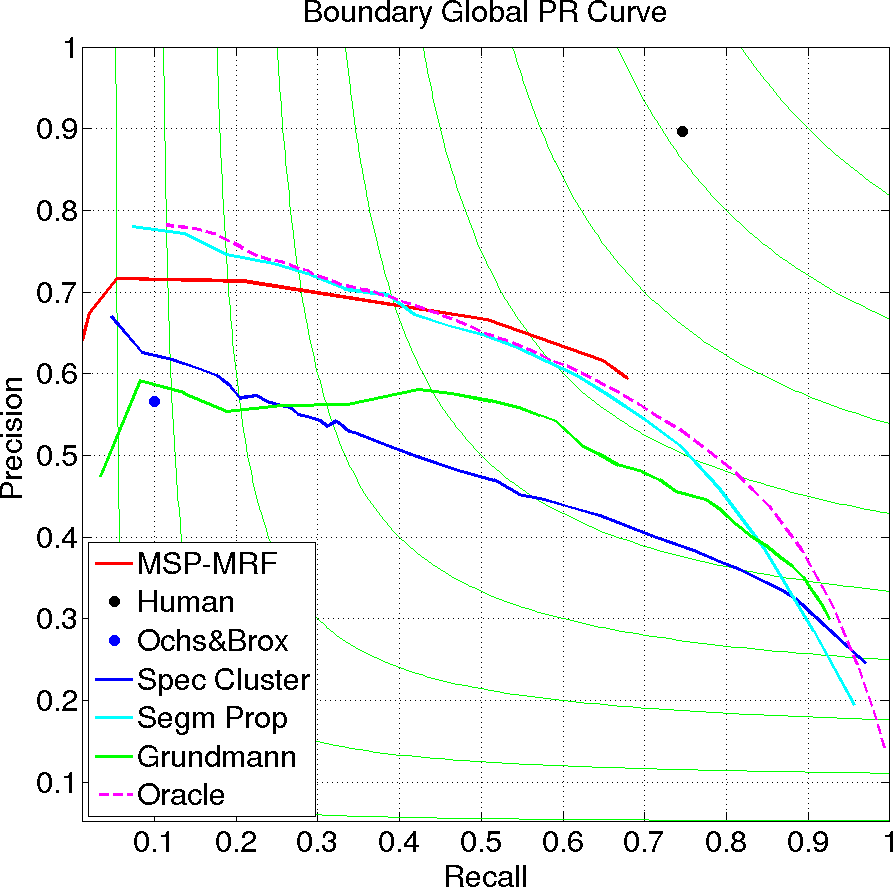}
		\end{subfigure}
		\begin{subfigure}{0.227\textwidth}
			\includegraphics[width=\textwidth]{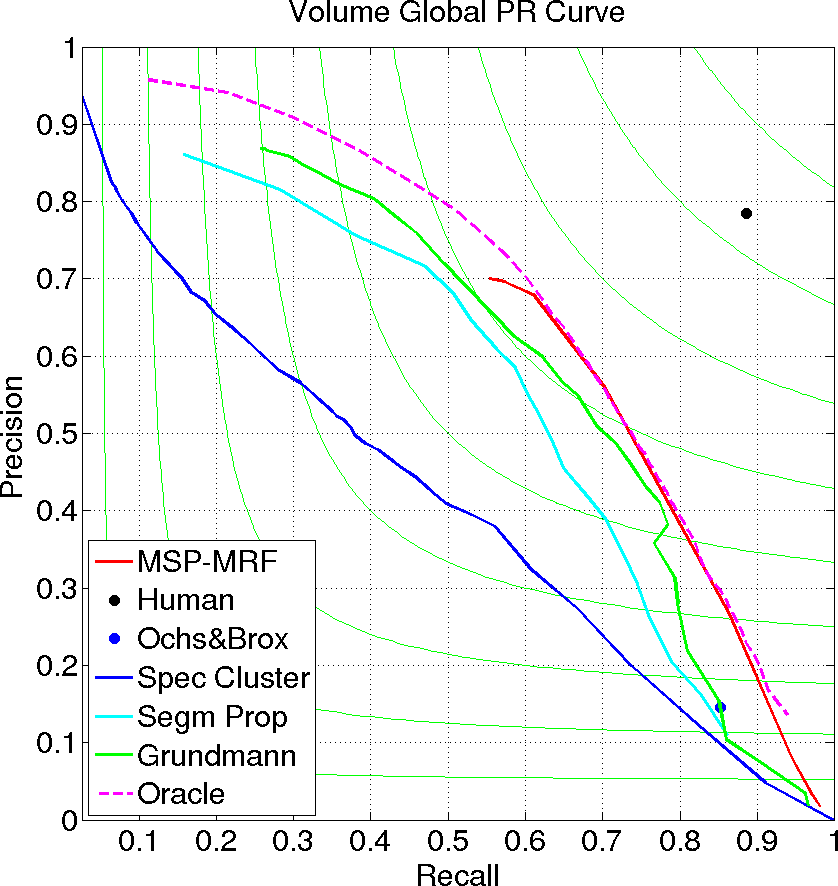}		
		\end{subfigure}
		\caption{PR curve comparison to other models.}
		\label{fig:res}
	\end{figure}
	
	\begin{figure}[!t]
		\centering
		\begin{subfigure}{0.245\textwidth}
			\includegraphics[width=\textwidth]{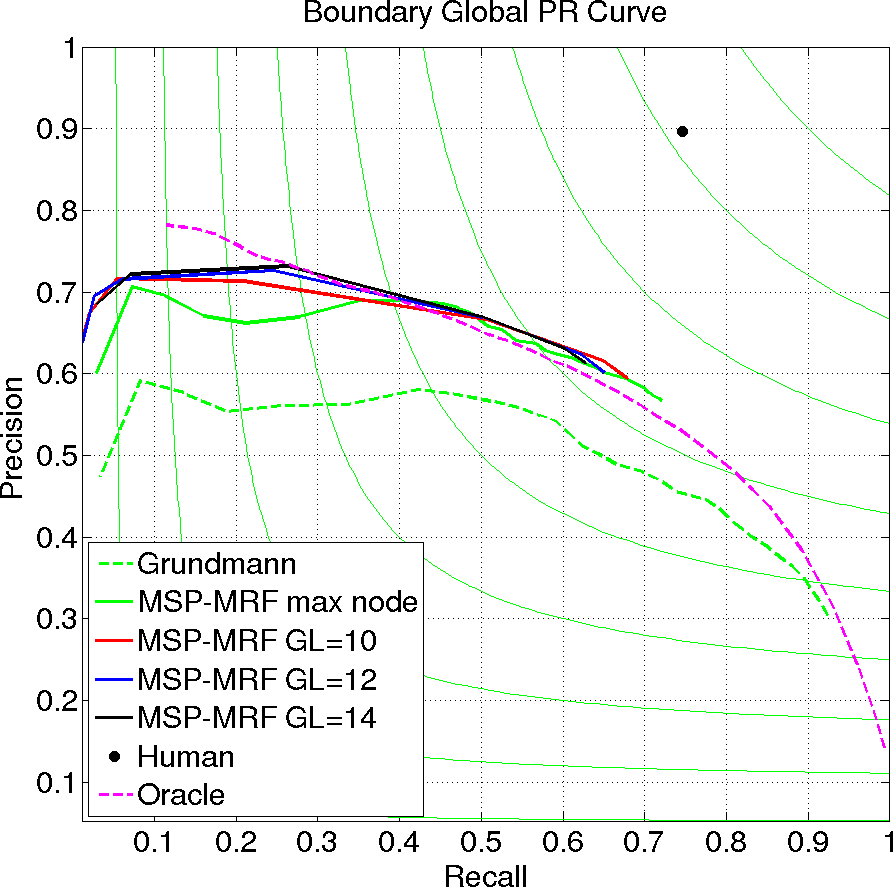}
		\end{subfigure}
		\begin{subfigure}{0.225\textwidth}
			\includegraphics[width=\textwidth]{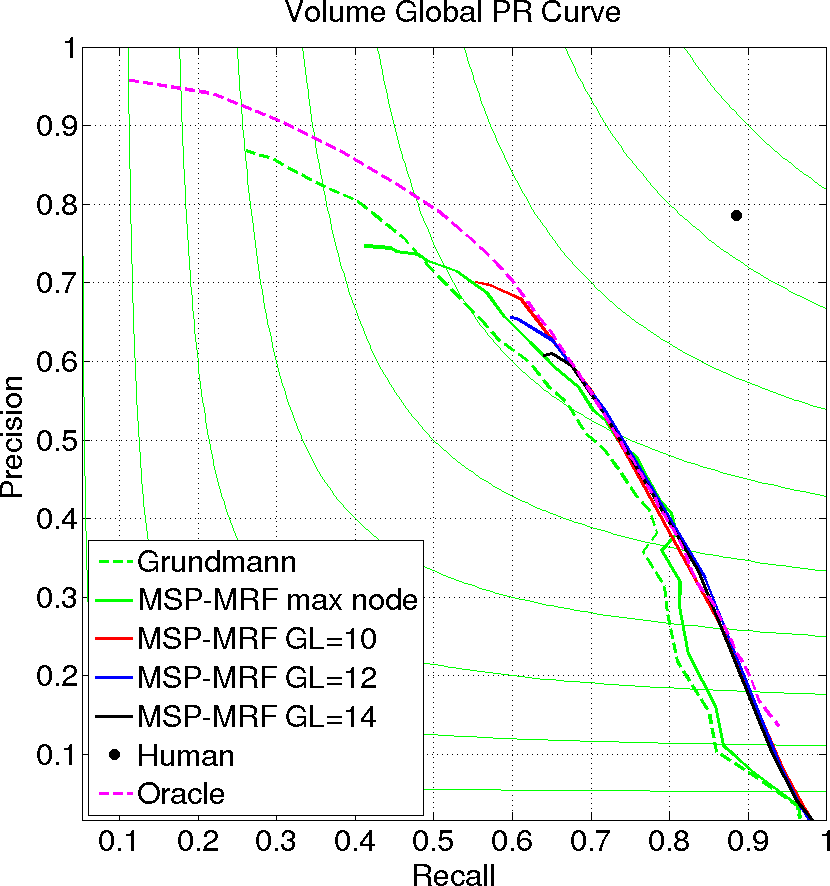}
		\end{subfigure}
		\caption{PR curve on different size of label set $\mathcal{L}$.}
		\label{fig:diff_gl}
	\end{figure}
	
	\begin{table*}[!htbp]
		\centering
		\begin{tabular}{| l || c |c| c||c|c|c||c|c|}
			\hline
			& \multicolumn{3}{|c||}{BPR} & \multicolumn{3}{|c||}{VPR} &    Length     &  NCL   \\ \hline
			Algorithm                                                    & ODS  & OSS  & AP           & ODS  & OSS  &      AP      & $\mu(\delta)$ & $\mu$  \\ \hline
			Human                                                        & 0.81 & 0.81 & 0.67         & 0.83 & 0.83 &     0.70     & 83.24(40.04)  & 11.90  \\ \hline\hline
			Ochs and Brox~\cite{Ochs11}                                  & 0.17 & 0.17 & 0.06         & 0.25 & 0.25 &     0.12     & 87.85(38.83)  &  3.73  \\
			Spectral Clustering~\cite{Galasso12}                         & 0.51 & 0.56 & 0.45         & 0.45 & 0.51 &     0.42     & 80.17(37.56)  &  8.00  \\
			Segmentation Propagation~\cite{Galasso13}                    & 0.61 & 0.65 & \textbf{0.59}         & 0.59 & 0.62 &     0.56     & 25.50(36.48)  & 258.05 \\
			$\mathcal{G}^Q\equiv\mathcal{G}'$ SC~\cite{Galasso14}        & 0.62 & 0.66 & 0.54         & 0.55 & 0.59 &     0.55     & 61.25(40.87)  & 80.00  \\
			$[\mathcal{M(G^{SC}}_2)]^\text{NCut}$-1SC~\cite{Galasso14G}  & 0.61 & 0.64 & 0.51         & 0.58 & 0.61 &     0.58     & 60.48(43.19)  & 50.00  \\
			Grundmann \etal~\cite{Grundmann10}                           & 0.57 & 0.62 & 0.48         & 0.61 & 0.65 &     0.61     & 51.83(39.91)  & 117.90 \\
			MSP-MRF                                                      & \textbf{0.63} & \textbf{0.67} & 0.57         & \textbf{0.65} & \textbf{0.67} &     				\textbf{0.64}     & 35.76(38.72)  & 168.93 \\ \hline\hline
			\textit{Oracle}~\cite{Galasso13}                              & 0.62 & 0.68 & 0.61         & 0.65 & 0.67 &     0.68     &       -       & 118.56 \\ \hline
		\end{tabular}
		\vspace{-0.5em}
		\caption{Performance of MSP-MRF model compared with state-of-the-art video segmentation algorithms on VSB100.}
		\label{tab:1}
	\end{table*}
	
	
	\subsection{PR Curve on High recall regions}
	\label{sec:gl}
	We specifically consider high recall regions of segmentation since we are typically interested in videos with relatively few objects.
	Our proposed method improves and rectifies state-of-the-art video segmentation of greedy agglomerative clustering~\cite{Grundmann10}, because we make use of structural information of object boundary, color, optical flow, texture and temporal correspondence from long trajectories.  Figure \ref{fig:res} shows that the proposed method achieves significant improvement over state-of-the-art algorithms.  MSP-MRF improves in both BPR and VPR scores such that it is close to {\it Oracle} which evaluates contour based superpixels on ground truth.  Hence, it is worth noting that {\it oracle} is the best accuracy that MSP-MRF could possibly achieve because MSP-MRF takes contour based superpixels from \cite{Arbelaez11} as well.
	
	The proposed MSP-MRF model rectifies agglomerative clustering by merging two different labels of vertices if it reduces overall cost defined in (\ref{eq:model}).  By increasing the number of edges in the graph by lowering threshold value, the model leads to coarser grained segmentation.  As a result, MSP-MRF only covers higher recall regions from precision-recall scores of the selected label set size $|\mathcal{L}|$ from \cite{Grundmann10}.  A hybrid model that covers high precision regions is described in Section \ref{sec:hybrid}.
	
	Figure \ref{fig:diff_gl} illustrates the PR curve of MSP-MRF on different granularity levels of label set $|\mathcal{L}|$ in node potential (\ref{eq:node}).  Dashed-green line is the result of greedy agglomerative clustering~\cite{Grundmann10}.  Solid-green line is the result of MSP-MRF with edge threshold $\tau$ set to $1$, which leaves no edge in the graph.  
	The figure shows that results of MSP-MRF are stable over different size of $|\mathcal{L}|$, particularly in the high recall regions.
	
	
	\subsection{Hybrid Model for Over Segmentation}
	\label{sec:hybrid}
	The proposed model effectively merges labels of each pair of nodes according to edge set $\mathcal{E}$.  As the number of edges increases, the size of the inferred label set will decrease from $|\mathcal{L}|$, which will cover higher recall regions.  Although we are interested in high recall regions, the model needs to be evaluated on high precision regions of PR curve.  For this purpose, we take a hybrid model that obtains rectified segmentation results from MSP-MRF on the high recall regions but retains segmentation result of \cite{Grundmann10} on high precision regions as an unrectified baseline. 
	
	Table \ref{tab:1} shows performance comparison to state-of-the-art video segmentation algorithms.  The proposed MSP-MRF model outperforms state-of-the-art algorithms on most of the evaluation metrics.  BPR and VPR is described in Section \ref{sec:dataset}.  Optimal dataset scale(ODS) aggregates F-scores on a single fixed scale of PR curve across all video sequences, while optimal segmentation scale(OSS) selects the best F-score with different scale for each video sequence.  All the evaluation metrics are followed from dataset~\cite{Galasso13}.  It is worth noting that our MSP-MRF model achieves best ODS and OSS results for both BPR and VPR evaluation measurements, which are equivalent to results of \textit{Oracle}.  As described in Section \ref{sec:gl}, \textit{Oracle} is a model that evaluates contour based superpixels on ground truth.  
	
	MSP-MRF infers segmentation label by integrating object boundary, global structure and temporal smoothness based on ~\cite{Grundmann10}.  The result shows that incorporating boundary and global structure rectifies ~\cite{Grundmann10} by significant margin.  It should be noted that result of ~\cite{Grundmann10} is higher than previously reported in ~\cite{Galasso13}.  We assume this is due to implementation updates on \cite{Grundmann10} over recent years.  Qualitatively, we observe that recent implementation of \cite{Grundmann10} detects objects whose appearance is less distinctive from background, where the previous implementation could not elucidate objects under those circumstances. 
	
	
	\section{Conclusion}
	\label{sec:conclusion}
	In this paper, we have presented a novel video segmentation model that considers three important aspects of video segmentation.  The model preserves object boundary by defining vertex set from contour based superpixels.  In addition, temporal smooth label is inferred by providing unary node potential from agglomerative clustering label likelihood.  Finally, global structure is enforced from pairwise edge potential on object boundary, color, optical flow motion, texture and long trajectory affinities.  Experimental evaluation shows that the proposed model outperforms state-of-the-art video segmentation algorithm on most of the metrics.
	
	
	{\small
		\bibliographystyle{ieee}
		\bibliography{egbib}
	}
	
\end{document}